\documentclass[preprint,5p,compress]{elsarticle}

\providecommand{\revisioncolor}[0]{}
\providecommand{\revision}[1]{{\revisioncolor #1}}

\usepackage{xcolor}
\usepackage{tikz}
\usepackage{amssymb}
\usepackage{amsmath}
\usepackage{booktabs}
\usepackage{makecell}

\usetikzlibrary{shapes.misc, positioning, shapes.multipart, arrows.meta}
\tikzset{
    layer/.style={
        draw, 
        rectangle split,
        minimum width=2cm,
        rectangle split parts=3, 
        rectangle split part fill={black!10,white,black!10},
        rounded corners,
        inner sep=5pt,
    }
}

\usepackage{hyperref}

\journal{Neurocomputing}
\begin{document}

\begin{frontmatter}

\title{\revision{An analysis} on the use of autoencoders for representation learning:\\ Fundamentals, learning task case studies, \revision{explainability and challenges}\tnoteref{t1}}

\author[UGR]{David Charte\corref{cor1}}
\ead{fdavidcl@ugr.es}

\author[UJA]{Francisco Charte}
\ead{fcharte@ujaen.es}

\author[UJA]{Mar\'ia J. del Jesus}
\ead{mjjesus@ujaen.es}

\author[UGR]{Francisco Herrera}
\ead{herrera@decsai.ugr.es}

\cortext[cor1]{Corresponding author.}
\tnotetext[t1]{This manuscript has been accepted at Neurocomputing. \textcopyright 2020. This version is made available under the CC-BY-NC-ND 4.0 license \url{http://creativecommons.org/licenses/by-nc-nd/4.0/}. The final authenticated version is available at \url{https://doi.org/10.1016/j.neucom.2020.04.057}.}
\address[UGR]{Department of Computer Science and A.I., University of Granada, 18071 Granada, Spain}
\address[UJA]{Department of Computer Science, University of Ja\'en, 23071 Ja\'en, Spain}

\begin{abstract}
In many machine learning tasks, learning a good representation of the data can be the key to building a well-performant solution. This is because most learning algorithms operate with the features in order to find models for the data. For instance, classification performance can improve if the data is mapped to a space where classes are easily separated, and regression can be facilitated by finding a manifold of data in the feature space. As a general rule, features are transformed by means of statistical methods such as principal component analysis, or manifold learning techniques such as Isomap or locally linear embedding. From a plethora of representation learning methods, one of the most versatile tools is the autoencoder. In this paper we aim to demonstrate how to influence its learned representations to achieve the desired learning behavior. To this end, we present a series of learning tasks: data embedding for visualization, image denoising, semantic hashing, detection of abnormal behaviors and instance generation. We model them from the representation learning perspective, following the state of the art methodologies in each field. A solution is proposed for each task employing autoencoders as the only learning method. The theoretical developments are put into practice using a selection of datasets for the different problems and implementing each solution, followed by a discussion of the results in each case study \revision{and a brief explanation of other six learning applications. We also explore the current challenges and approaches to explainability in the context of autoencoders.} All of this helps conclude that, thanks to alterations in their structure as well as their objective function, autoencoders may be the core of a possible solution to many problems which can be modeled as a transformation of the feature space.

\end{abstract}


\end{frontmatter}


\section{Introduction}

Creating new representations of data is a fundamental task in most machine learning tasks. \revision{First off, certain types of problems that require a classifier or a regressor will certainly benefit from transformations of the features which facilitate their work~\cite{domingos-useful}. In addition to this, there exists a variety of problems whose solution relies strongly on finding an appropriate representation of the data. Although the use of representation learning techniques is mainly used as a complement to other learners in the former case, in the latter one these methods become the focus. This work highlights some of these situations, with specific applications that can be modeled as representation learning problems.}

\revision{The features that are used as input conform one of the most important factors when building machine learning models.} When the training set contains intact data from its collection or measurements, it may not be ready for treatment yet. Instead, it is common for data to be expressed with redundant or uninformative variables and for it to include some level of noise. These and other obstacles presented by the data~\cite{lorena2019complex} are the reason why most of the manual work of building machine learning models is spent in the preprocessing stage~\cite{garcia2015data}. 

The success of a classifier, a regressor or other models will greatly depend on the quality of the features it can learn from. \revision{For instance, decision trees, regardless of whether the task is  classification or regression, attempt to find the most informative variables to branch at each step~\cite{kotsiantis2007supervised}; support vector machines calculate the hyperplane that best separates classes in a feature space originating from specific transformations of the original one~\cite{kotsiantis2007supervised}, and k-means clustering computes distances among pairs of instances and thus depends strongly on the input domain~\cite{Jain1999DataCA}.}
\revision{As a result, it is of vital importance that the features provided to these learners are useful and as independent as possible.}

However, finding alternative representations for data is not only a medium to build classification and regression models, but it may be an end in itself in many applications. For example, finding compact binary codes that represent text documents~\cite{salakhutdinov}, compressing signals to a lower resolution without losing information~\cite{compression}, transforming the problem domain to a different one~\cite{transferlearning}, or producing filtered versions of images with less distortions~\cite{xie}.

Learning representations usually consists in feature engineering~\cite{domingos-useful} or feature extraction~\cite{featext}, depending on whether new features are computed manually by human intervention (either by selection~\cite{dash1997feature} or simple arithmetic operations) or they are generated, evaluated and selected by the machine. Feature engineering leverages expert knowledge and human creativity in order to select features and operate with them in a way that results in a new feature set which seems appropriate for predictors to work with. Nowadays there exist many automatic approaches to feature learning, which relieve users from the tedious task of engineering new features~\cite{bengio}. These methods range from probabilistic to topological and from shallow to deep: principal component analysis~\cite{PCABook}, Isomap~\cite{Isomap}, locally linear embedding~\cite{LLE} and Laplacian eigenmaps~\cite{LaplacianEigenmaps}, among others.

With the introduction of deep neural networks, the representation learning stage became integrated within the predictors themselves~\cite{lecun-dl}. These techniques iteratively optimize the classification performance by modifying the weights in several layers of individual neurons which compute a hierarchy of abstractions over the original data. For this purpose, the backpropagation algorithm~\cite{rumelhart1988learning} allows to efficiently accumulate gradients along the network, so that an optimizer such as Stochastic Gradient Descent~\cite{robbins1951stochastic} or one of its derivatives~\cite{duchi2011adaptive,zeiler2012adadelta,kingma2015adam,tieleman2012lecture} may compute each weight update. Since neural networks can be structured as needed for each kind of problem, they are able to function as standalone feature learners as well. This is the case of autoencoders (AEs)~\cite{charte-tutorial}, neural architectures whose objective is to find the best representation for the data according to the criterium defined by their loss function.

The objective of this paper is to analyze how AEs can serve as the main basis for solving a wide variety of learning tasks and demonstrate this with concrete applications and experimental results. 
Throughout the paper, we examine several case studies that expose the adaptability of AEs to these problems.\begin{itemize}
    \item First, an example of data embedding onto a very low dimensional space for visualization and exploratory analysis. 
    \item Then, a case where noisy signals are to be repaired by the model.
    \item Later, a different example where very high dimensional sparse data, such as text documents, is to be compressed onto compact binary codes in a semantic way.
    \item Additionally, we study anomaly detection, the situation where abnormal patterns are to be detected in sequences but no anomalies are available to learn from.
    \item As a last case study, we propose the generation of new instances which do not belong to the training set. 
\end{itemize}    

Other applications are also briefly discussed: image superresolution, image compression, transfer learning, human pose recovery and recommender systems. 

As a starting point, we provide the reader with the necessary background knowledge about the field of representation learning, as well as a summary of the main features of AEs that make them a good candidate model to solve the different problems later approached. The solutions to these tasks using AEs as the only automatic learner highlight their potential and flexibility as feature extraction techniques. 

\revision{Following the current increase in search for developing explainable models \cite{arrieta2020explainable}, the main approaches for obtaining interpretable predictions are summarized, finding that quality features can be the key to explainable solutions. AE models which can build helpful features are also highlighted.}

The rest of this paper is structured as follows. Section~\ref{sec:bg} describes the background of the problems and techniques above introduced. Section~\ref{sec:ae} details the inner workings of AEs. Section~\ref{sec:cs} further develops on several case studies where AEs resolve feature learning tasks and outlines other existing learning applications, \revision{and Section~\ref{sec:ch} describes the current state of the art in explainable AI and how AEs are involved}. Lastly, Section~\ref{sec:co} concludes the text.


\section{Background: \revision{feature learning and deep representation learning}}\label{sec:bg}

\revision{This section 
explains some well-known methods that can extract features from data.} Afterwards, it introduces deep learning techniques.

\subsection{Classical feature learning methods}

Traditionally, feature extraction methods have been developed with linear as well as nonlinear transformations of the variables~\cite{featext}. They can be considered nonconvex or convex, according to whether the objective function presents local optima or not, respectively~\cite{DimRecComparative}. Many of these techniques perform unsupervised learning, but others are supervised~\cite{FisherLDA,lds,adv} or even semi-supervised~\cite{lsss}. Next, a summary of typical feature learning methods is provided.

\paragraph{Linear methods} The most common linear feature extraction methods are the following. Principal component analysis (PCA) consists in extracting successive variables or \textit{principal components} with maximum variance while being uncorrelated with the previous components. It is a statistical technique developed geometrically by Pearson~\cite{PCA} and algebraically by Hotelling~\cite{PCAHotelling}, but an analytical derivation can be found in~\cite{PCABook}. Factor analysis~\cite{PCAandFA} is a similar procedure to PCA which considers a set of latent variables or \textit{factors} that are not observed but are linearly combined to produce the final variables. Linear discriminant analysis~\cite{FisherLDA} is a supervised statistical technique which attempts to find linear combinations of features to project samples onto new coordinates that best discriminate classes, albeit making some assumptions about the distribution of the data. 

\paragraph{Nonlinear methods} Some well known nonlinear approaches to feature extraction are kernel PCA, restricted Boltzmann machines and manifold learning methods. Kernel PCA~\cite{KernelPCA} extends PCA to nonlinear combinations of features by projecting samples onto higher-dimensional spaces and using the kernel trick~\cite{kernelmethods}. Restricted Boltzmann machines are undirected graphical probabilistic models, also known as harmoniums~\cite{Harmonium}, with one visible layer and one hidden layer that acts as the set of extracted features. They can be trained using the contrastive divergence algorithm~\cite{ContrastiveDivergence}. Many nonlinear feature learning methods attempt to find coordinates for a lower dimensional structure embedded in the original features, namely, a manifold. Multidimensional scaling (MDS) is one of the first techniques that can be considered manifold learning, as its objective is projecting samples in a low-dimensional space while translating as much information of pairwise distances as possible. There are several variants of MDS, one of them is Sammon mapping~\cite{Sammon}, which improves on MDS by using a different cost function which stresses large distances similarly to small ones. Isomap~\cite{Isomap} is a more recent extension of MDS which looks for the coordinates that describe the actual degrees of freedom of the data while preserving distances among neighbors and geodesic distances (the length of the shortest path that connects two points in the manifold). Locally Linear Embedding~\cite{LLE} also seeks a manifold which preserves neighbors but, in order to maintain the local structure, it linearly reconstructs each point from its neighbors. Laplacian eigenmaps~\cite{LaplacianEigenmaps} is a procedure that builds a graph based on the neighborhood structure of the data, and from it a weight matrix whose eigenvectors can be used to compute new coordinates for each point.

\subsection{Deep representation learning}

Deep learning architectures are hierarchies of abstractions of the input feature space and, as such, they compute several transformations of the features before reaching a response. In some cases, these can be seen as learned representations, since they must be able to capture the relevant information from each instance in order to output an accurate result. This effect can be observed especially in convolutional neural network classifiers, which are usually split into a feature extraction component formed by convolutional layers and a decision module composed by fully connected layers~\cite{imagenet}. Apart from neural networks with other objectives such as supervised classification or regression, there have been different approaches to shallow as well as deep neural structures for unsupervised feature learning~\cite{reviewdl}, such as self-organizing Kohonen maps~\cite{kohonen1990self,koikkalainen1990self}, predictability minimization~\cite{schmidhuber1996semilinear}, restricted Boltzmann machines~\cite{DLBookRBM}, \revision{deep belief networks~\cite{hinton,dbngics,improvedpart} }and AEs~\cite{kramer1991nonlinear,oja1991}. There have been many instances of these unsupervised techniques being used to either pre-train or provide feature transformations for supervised models~\cite{bengio2012deep}.

AEs are probably the most versatile unsupervised neural network models. They essentially combine some kind of bottleneck or restriction in the learned data representations with the objective of reconstructing and repairing the original input from that representation~\cite{charte-tutorial}. There are several ways to impose restrictions that produce interesting representations, and the reconstruction objective will cause the network to retain all invariant feature information along its weights, so that the representation or encoding holds mainly instance-specific traits. For example, undercomplete AEs project inputs into lower-dimensional encodings, sparse AEs obtain representations with very few activated neurons, and denoising AEs attempt to repair partially corrupted data.

Their versatility is demonstrated by the amount of applications AEs have and their diversity. Across the rest of this work, we focus on certain applications of representation learning that are solved with AEs and we analyze how each model is built and trained.

\section{Autoencoder fundamentals}\label{sec:ae}

AEs are neural network structures designed with the purpose of learning new features. Throughout the following subsections, their main characteristics and differentiating aspects are outlined, and some ways to influence the encoded variables are discussed.

\subsection{Origin and essentials of autoencoders}

AEs were originally conceieved as a way of initializing neural networks~\cite{ballard} and continued fulfilling that purpose for some time, serving as a starting point for training of deep networks as well~\cite{hinton2006fast}. Over the last years, other applications for AEs have been emerging and at the same time other approaches to neural network training and regularization have succeeded over AEs~\cite{relu,dropout}. As a consequence,  the common uses for AEs have shifted from helping train other neural networks to other applications of their own.

In general, the training process required to learn an AE can be unsupervised, that is, it does not need labels or class information in order to generate a model for the data. Instead, it extracts useful information from each instance by feeding its feature vector through some transformations which impose a bottleneck or restriction on the possible representations it can compute. Then, the representation is mapped to the original feature space through a similar set of transformations, and the AE is evaluated according to the fidelity of the reconstruction. This feedback allows to modify the parameters iteratively until convergence is reached.

\begin{figure}[ht]
    \centering
    \includegraphics[width=\linewidth]{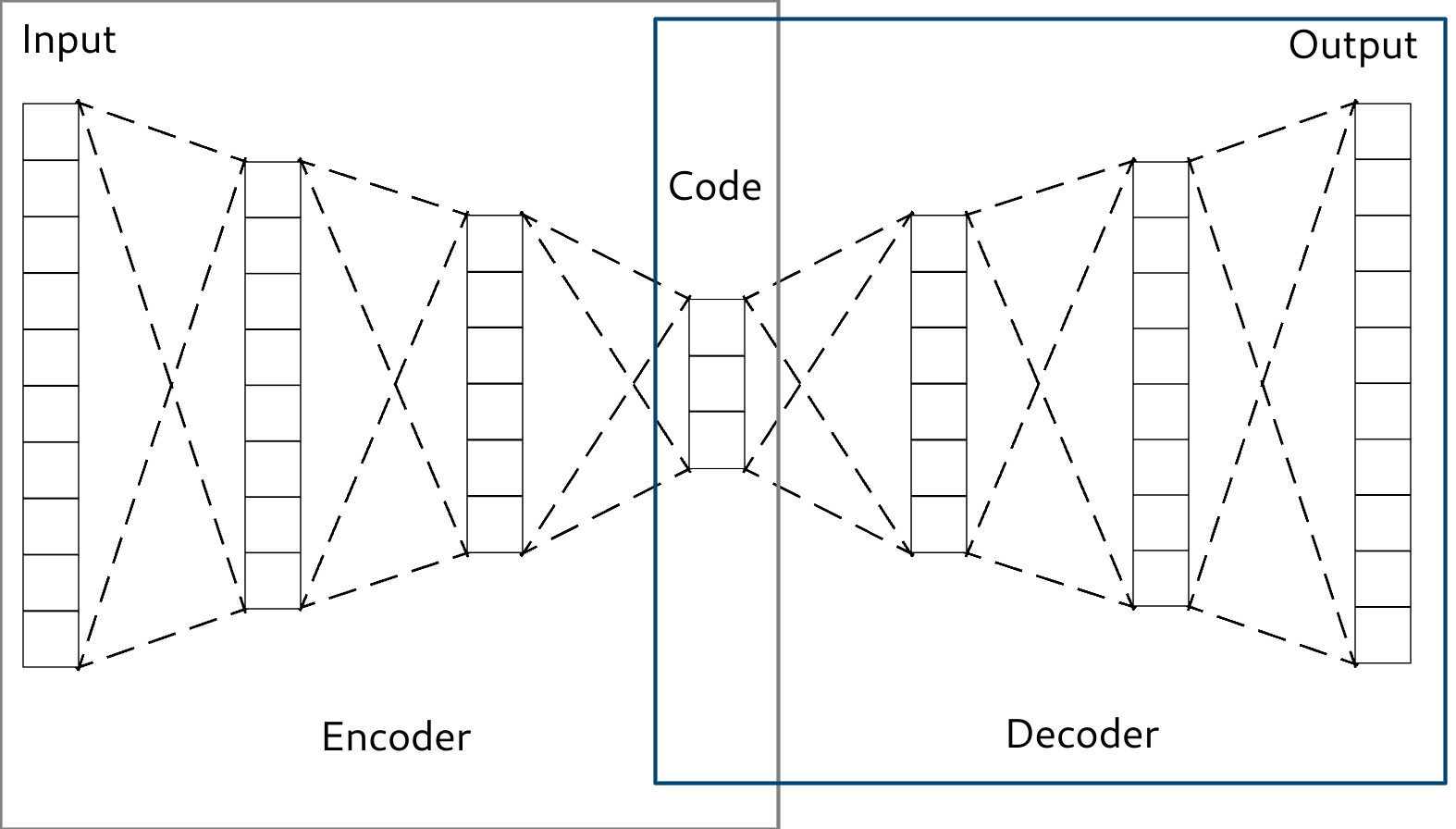}
    \caption{Illustration of the general structure of a basic AE: an encoder and a decoder connected by the encoding layer}
    \label{fig:ae}
\end{figure}

AEs take the form of a neural network with at least one hidden layer and two components, an encoder and a decoder, which are connected by the coding layer~\cite{charte-tutorial}. These components are usually symmetric in layer shapes to each other, especially if they are implemented as fully connected neural networks. In certain occasions, even the weights of each layer in the decoder are tied to the corresponding layer in the encoder. In general terms, however, it suffices with the input layer of the encoding and the output layer having the same shape. Fig.~\ref{fig:ae} shows how the architecture of an AE may look like.

\begin{table}[ht]\centering
    \revision{\begin{tabular}{lp{.7\linewidth}}
        \toprule
        Symbol & Interpretation \\
        \midrule
        $\theta$ & Full set of parameters of the AE (weights and biases) \\
        $\mathcal X$ & Set of input instances \\
        $\mathcal Z$ & Set of instances in encoding space \\
        $n$ & Dimension of input space \\
        $k$ & Dimension of encoding space \\
        $f$ & Encoder mapping \\
        $g$ & Decoder mapping \\
        $d$ & Distance function in input space \\
        $r$ & Regularization function \\
        \bottomrule
    \end{tabular}}
    \caption{\revision{\label{tbl:symbols}Intepretation of symbols used in the formulae}}
\end{table}

In summary, an AE can be seen as the composition of an encoding map $f$ which projects inputs onto a different feature space, and a decoding map $g$ which operates inversely \revision{(see Table~\ref{tbl:symbols} for the meaning of all symbols used below)}. The main objective of the AE is to recover as much information as possible of the original input, so it will attempt to minimize a distance between the inputs and the outputs:

\begin{equation}\label{eq:objective}
\min_{\theta} \sum_{x\mathbin{\revision{\in\mathcal X}}}d(x, g_{\theta}(f_{\theta}(x)))
\end{equation}

The distance function $d$ used in the loss function is usually either the mean squared error, see Eq.~\eqref{eq:mse}, or the cross entropy, shown in  Eq.~\eqref{eq:xent}. In the first case, data may not be normalized and the output units should use an unbounded activation function. For a cross entropy loss, each input and output variable is modeled as following a Bernoulli distribution, so data should be scaled to the $[0,1]$ interval and output units could make use of a sigmoid activation.

The mean squared error for an input $x$ and output $x'$ of length $n$ is defined as:
\begin{equation}\label{eq:mse}
    d(x, x') = \frac 1 n\sum_{i=1}^n(x_i - x'_i)^2
\end{equation}
Similarly, the binary cross entropy for the same input and output is computed as:
\begin{equation}\label{eq:xent}
    d(x, x') = - (x \bullet \log(x') + (1-x)\bullet \log(1-x')),
\end{equation}
where $\bullet$ denotes element-wise product and all other operations are also performed element-wise.

\subsection{Modeling the coding layer}\label{sec:modelcode}

The main objective of the AE (Eq.~\ref{eq:objective}) only promotes faithful reconstructions without explicitly considering any aspect about the codes used. This can be enough in many cases where the codes are low dimensional and they can capture only the relevant information of the instances just by training to reconstruct accurately. Notwithstanding, there are situations that require considering a more general case of the objective, which allows penalizing certain behaviors of the encoding found by the network, or even the values of the parameters themselves (Eq.~\ref{eq:penalty}).

\begin{equation}\label{eq:penalty}
    \min_{\theta}\sum_{x\in\mathcal X} d(x,g_{\theta}(f_{\theta}(x)))+r_1(f_{\theta}(\mathcal X))+r_2(\theta)
\end{equation}

A straightforward example of this kind of restrictions is the sparse AE~\cite{lee_sparse_2008,ng2011sparse}, which adds a penalty for high activation rates in the neurons of the code layer (Eqs.~\ref{eq:sparse1} and \ref{eq:sparse2}):
\begin{align}
    \label{eq:sparse1}
    r(\mathcal Z)&=\sum_{j=1}^k(\rho-\rho_j)^2\mbox{, or}\\
    \label{eq:sparse2}
    r(\mathcal Z)&=\sum_{j=1}^k \rho \log\frac{\rho}{\rho_j}+(1-\rho)\log\frac{1-\rho}{1-\rho_j},
\end{align}
where $\rho_j=\frac{1}{\lvert\mathcal Z\rvert}\sum_{z\in \mathcal Z}z_{\revision{j}}$ is the average activation vector, $k$ is the length of the code and $\rho$ is the desired activation rate.

Other, more sophisticated variations on the AE with different penalties are the contractive AE~\cite{contractive,rifai2011higher}, which promotes finding and preserving any local structure from the original feature space, and the variational AE~\cite{variational}, which uses a penalty to impose a distribution to the codes computed by the encoder.

Penalties on the codes are not, however, the only way of incentivizing a behavior on the encoder mapping. Denoising AEs~\cite{vincent2008extracting,sdae} establish a slightly different criterion to evaluate the performance of the reconstruction: the network must be able to repair any noise or corruption from the input.  Robust AEs~\cite{qi_robust_2014} use another objective function, correntropy~\cite{correntropy}, which has a similar effect in repairing several kinds of noise from the input data.

\revision{\subsection{Evaluation metrics}}

\revision{The quality of learned features can be evaluated by the model's ability to project instances back to the original feature space. For this purpose, regression metrics can be used. Some common metrics which serve to assess the usefulness of the learned features are the following, where $x$ is the original feature vector and $x'$ is the reconstruction, mapped from the encoding space back onto the input space:
\begin{itemize}
    \item Mean squared error (Eq.~\ref{eq:mse}) and root mean squared error: \[\operatorname{RMSE}(x,x')=\sqrt{\frac 1 n \sum_{i=1}^n \left( x_i - x'_i\right)^2}\]
    \item Mean absolute error: \[\operatorname{MAE}(x,x')=\frac 1 n \sum_{i=1}^n \left\lvert x_i - x'_i\right\rvert\]
    \item Mean absolute percentage error \[\operatorname{MAPE}(x,x')=\frac 1 n \sum_{i=1}^n \left\lvert \frac{x_i - x'_i}{x_i}\right\rvert\]
\end{itemize}
In certain cases, the encoded features can also be evaluated independently from the original features, by assessing their quality with respect to their complexity, class separability and overlap \cite{lorena2019complex}. This usually requires that data belongs to a classification problem so that a class is defined for each instance. }

\subsection{Beyond unsupervised autoencoders}

Although the objective of an AE usually does not involve direct prediction of labels, it can sometimes learn from classified examples. The most straightforward way to introduce class information into the AE is to modify the loss function so it propagates different errors according to the class of each instance. For example, we could weight each class differently. Assuming the classes are binary, \revision{dividing the dataset into $\mathcal X^+$ for positive instances and $\mathcal X^-$ for negative ones, and $\alpha$ is a parameter in $[0,1]$}, the objective in Eq.~\eqref{eq:weight}
\begin{equation}\label{eq:weight}
    \min_{\theta}(1-\alpha)\sum_{x\in \mathcal X^-} d(x,g_{\theta}(f_{\theta}(x)))+\alpha\sum_{x\in \mathcal X^+} d(x,g_{\theta}(f_{\theta}(x)))
\end{equation}
would give more importance to reconstructing one of the classes, which may help if the aim is to find a manifold for that class and the other one is less relevant.

Several uses of label information can be found in the proposal of the adversarial AE~\cite{adv}. This AE has a similar behavior to the variational AE in that it also forces the codes to follow a given distribution. Instead of using just a loss penalty, it adds a generator which samples the distribution, and a discriminator which attempts to distinguish distribution samples from codes belonging to actual instances, analogous to a generative adversarial network~\cite{gan}. The label information can be used then to locate each label in a region of the distribution, by feeding labels as well as codes to the discriminator. Alternatively, labels can be feeded to the decoder, which causes the codes to discard label information and instead model style in the data.

Another step forward in introducing label information in AEs would be for them to be able to predict labels as well. Some work has been already done along these lines, by training an encoder and decoder simultaneously to reconstruct and to produce codes as similar as possible to the labels in a one-hot format~\cite{zhuang2015supervised}.

\section{Learning task case studies}\label{sec:cs}

The following subsections detail several real examples of application of AEs: embedding data onto a very low-dimensional space for visualization purposes, reducing the noise in images, computing semantic hashes for large text documents, finding anomalous behaviors in sequences and generating new instances outside the training set. For each application, a relevant dataset has been selected and a model has been specifically designed to solve the problem. The basic traits of all chosen datasets can be found in Table~\ref{tbl:datasets}.

\begin{table}[hb]
    \centering
    \resizebox{\linewidth}{!}{
        \begin{tabular}{llrrr}
            \toprule
            \makecell[l]{Dataset} & \makecell[l]{Application} & \makecell[r]{Input\\features} & \makecell[r]{Training\\examples} & \makecell[r]{Test\\examples}\\\midrule
            CPU Activity & Visualization  & 21  & 6553 & 1639\\
            Satellite image & Visualization  & 36  & 5142 & 1288\\
            STL10~\cite{stl10} & Noise reduction & $96\times 96\times 3$  &  5000 & 8000  \\
            Bibtex\cite{bibtex} & Semantic hashing & 1836 & 5916 & 1479 \\
            UNSW-NB15\cite{unsw} & Anomaly detection & 187 & 37000 & 175341 \\
            AT\&T faces & Instance generation & $64\times 64$ & 400 & - \\
            \bottomrule
            \end{tabular}
    }
    \caption{\label{tbl:datasets}Main traits of datasets used for the experiments}
\end{table}

\begin{figure}[ht]
    \centering\small
     \resizebox{\linewidth}{!}{
\begin{tikzpicture}[node distance=0.5cm]
\node[layer] (x) {\textsc{purpose: input}
\nodepart{second} Layer type: input data
\nodepart{third} Output shape: 1000};
\node[layer, right=3.2cm of x] (h1) {\textsc{encoding}
\nodepart{second} Dense
\nodepart{third} 10};
\node[layer, right=of h1] (h3) {\textsc{output}
\nodepart{second} Dense
\nodepart{third} 1000};
\path (x) -- node[sloped] (text) {forward direction} (h1);
\draw[-Latex] (x.east) -- (text) -- (h1.west);
\draw[-Latex] (h1.east) to (h3.west);
\end{tikzpicture}
}
    \caption{Example AE architecture. Each block represents a layer and is splitted into three parts: the meaning or purpose of the layer, the type of operation performed and its output shape (size of each dimension).}
    \label{fig:exampleae}
\end{figure}
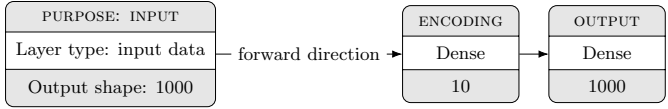

The models described below are each associated to a diagram describing the layer structure of the corresponding AE and the purpose of each layer. Please refer to Fig.~\ref{fig:exampleae} for an example of how each model is detailed.

All examples have been implemented and executed employing the following setup: Tensorflow~\cite{tensorflow} 1.14.0 and Keras~\cite{keras} 2.2.4 on top of Python 3.7 and R 3.6, running on an Intel Core i5-8400 CPU and a NVIDIA GeForce RTX 2060 GPU. The associated software can be found at the following GitHub repository: \url{https://github.com/ari-dasci/autoencoder-case-studies/}.

\subsection{Data visualization}

Most of the data collected nowadays, either from industries or from the web, is high-dimensional. Visualization techniques can help its interpretability, but the data generally needs to be summarized for this purpose. Traditionally, an alternative representation would be a subset of its features or its principal components~\cite{jolliffe-pca}. An AE, however, is able to automatically compute a representation that fits each dataset. This representation can be 2 or 3-dimensional if the AE is configured conveniently~\cite{yu2013embedding}, or if another embedding technique (such as t-SNE~\cite{tsne}) is used after a higher-dimensional encoding.

In particular, if our dataset consists of instances $(x, y)$ where $x$ is a feature vector and $y$ is its associated label, we can use a training subset to learn an autoencoder model with an encoding $f: \mathbb R^n\rightarrow \mathbb R^2$ resulting of the composition of the hidden layers up to the code layer. Then, encoded examples can be colored in a scatter plot according to their class.

Although a simple AE could fulfill the embedding task, it can be convenient to restrict or modify its behavior so as to influence the projection to the embedding space, in a way that improves how the populated regions in the original space are modeled. Along these lines, there are several approaches: denoising criteria~\cite{vincent2008extracting}, contractive regularizations~\cite{contractive} and embedding regularizations~\cite{yu2013embedding}.

\paragraph{Denoising criterion} A denoising AE~\cite{vincent2008extracting} trains, as briefly explained in Section~\ref{sec:modelcode}, by reconstructing partially corrupted inputs. In order to do this, a corruption or noise function introduces alterations on the input data: for example, a Gaussian noise $\xi\sim N(0,\sigma)$ would be used to produce the input $\revision{\nu(x)} = x + \xi$. The reconstruction error is now computed as $\sum_{x\revision{\in\mathcal X}} d(x, g(f(\revision{\nu(x)})))$. During the training process, the AE is forced to distinguish useful information from mere perturbations of the data. If the instances lie on a manifold in the original feature space, this can effectively train the AE to ``push back'' instances to the manifold by discarding small displacements from it. This can remove noise in the inputs as well as reconstruct some missing values if inputs are just an estimation~\cite{xie,li2015feature}. As a result, the encoding can serve as a set of coordinates for the manifold.

\paragraph{Contractive regularization} The contractive AE~\cite{contractive} uses an additional penalty in the training objective which promotes local invariance to displacements in many directions around the training samples, i.e., it is less sensitive to small perturbations especially in directions that lead outside the manifold. The penalty consists in the squared Frobenius norm of the Jacobian matrix of the encoder, that is, the sum of the squares of all first-order partial derivatives applied to all inputs: $\sum_x\lVert J_f(x)\rVert^2$. This can be seen as a generalization of L2 weight decay to the case where the encoder is nonlinear. This regularization favors encodings where all dimensions are contracted, but the reconstruction error prevents the AE from contracting dimensions along the manifold.

\paragraph{Embedding regularization} An alternative objective function for AEs can be the same loss function from other embedding techniques. This is the idea behind embeddings with AE regularization~\cite{yu2013embedding}, which combines the reconstruction error with one of several possible embedding loss functions coming from Laplacian eigenmaps~\cite{LaplacianEigenmaps}, multidimensional scaling~\cite{MDS} and margin-based embedding~\cite{hadsell2006dimensionality}. These loss functions evaluate the embedding by taking pairs of instances, and the AE is adapted the same way, by computing the embedding loss across all pairs of instances and the reconstruction loss across all instances. \\

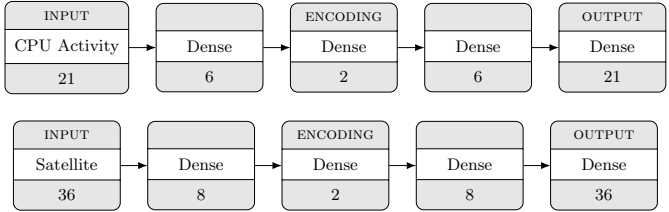
\begin{figure}[ht]
    \centering\small
    \resizebox{\linewidth}{!}{

        \begin{tikzpicture}[node distance=0.5cm]
            \node[layer] (x) {\textsc{input}
            \nodepart{second} CPU Activity
            \nodepart{third} 21};
            \node[layer, right=of x] (h1) {\textsc{}
            \nodepart{second} Dense
            \nodepart{third} 6};
            \node[layer, right=of h1] (h3) {\textsc{encoding}
            \nodepart{second} Dense
            \nodepart{third} 2};
            \node[layer, right=of h3] (h4) {\textsc{}
            \nodepart{second} Dense
            \nodepart{third} 6};
            \node[layer, right=of h4] (y) {\textsc{output}
            \nodepart{second} Dense
            \nodepart{third} 21};
            \draw[-Latex] (x.east) to (h1.west);
            \draw[-Latex] (h1.east) to (h3.west);
            \draw[-Latex] (h3.east) to (h4.west);
            \draw[-Latex] (h4.east) to (y.west);

            \node[layer, below=of x] (xx) {\textsc{input}
            \nodepart{second} Satellite
            \nodepart{third} 36};
            \node[layer, right=of xx] (z1) {\textsc{}
            \nodepart{second} Dense
            \nodepart{third} 8};
            \node[layer, right=of z1] (z3) {\textsc{encoding}
            \nodepart{second} Dense
            \nodepart{third} 2};
            \node[layer, right=of z3] (z4) {\textsc{}
            \nodepart{second} Dense
            \nodepart{third} 8};
            \node[layer, right=of z4] (yy) {\textsc{output}
            \nodepart{second} Dense
            \nodepart{third} 36};
            \draw[-Latex] (xx.east) to (z1.west);
            \draw[-Latex] (z1.east) to (z3.west);
            \draw[-Latex] (z3.east) to (z4.west);
            \draw[-Latex] (z4.east) to (yy.west);
            \end{tikzpicture}        
    }
    \resizebox{\linewidth}{!}{
    \begin{tikzpicture}[node distance=0.5cm]
        \end{tikzpicture}
    }
            
    \caption{AE architectures for visualization}
    \label{fig:visualizer}
\end{figure}

\begin{figure}[ht!]
    \centering
\includegraphics[width=\linewidth]{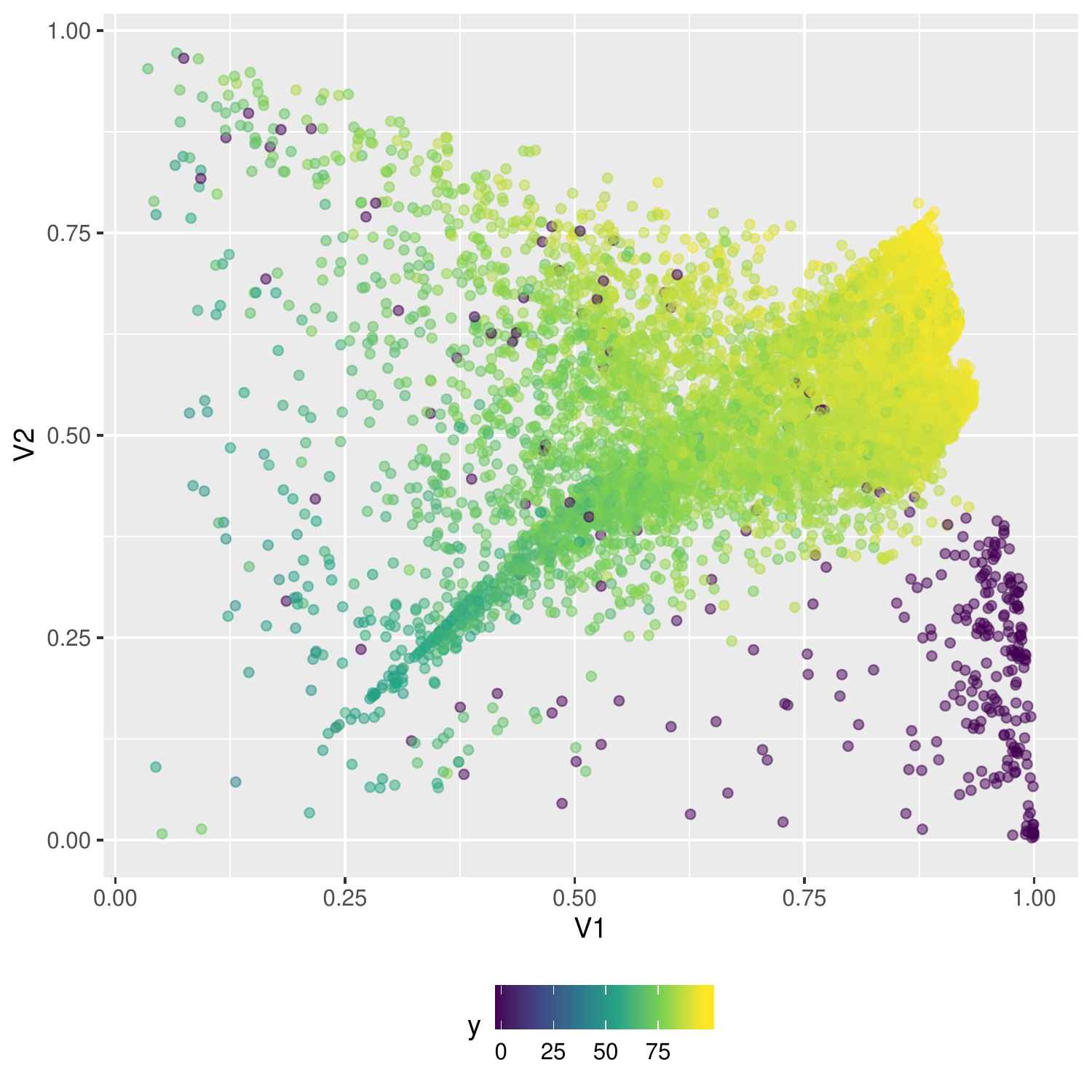} %
\includegraphics[width=\linewidth]{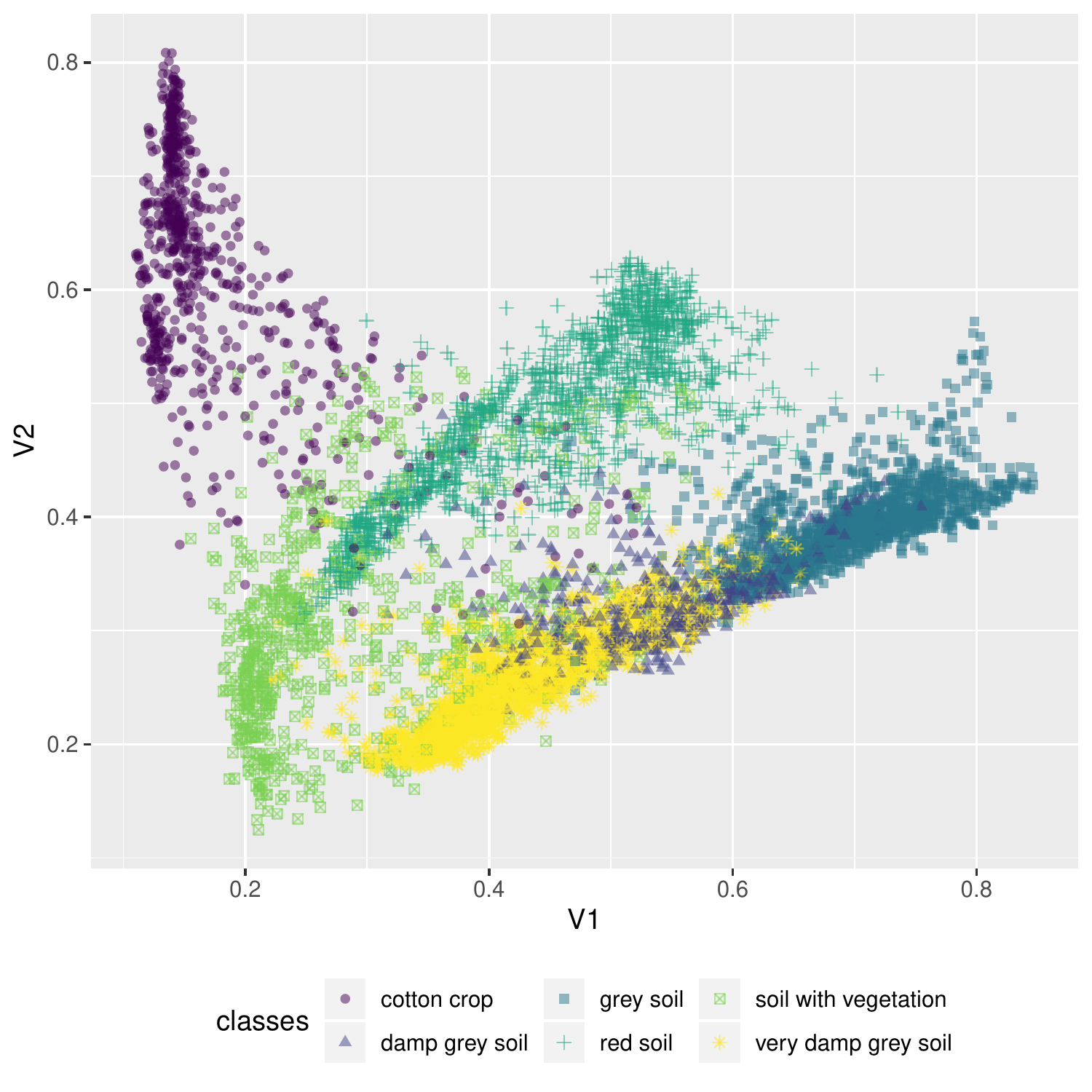}
    \caption{Embeddings learned by an unsupervised contractive AE. The top image shows the projection of the CPU Activity dataset where each point has been shaded according to the level of user activity. The bottom image displays the projected samples of the Satellite Image dataset, each one colored according to its class. }
    \label{fig:visualizations}
\end{figure}

For the purposes of demonstrating the capacity of AEs to find manifolds and appropriate embeddings, we have selected a regression dataset, CPU activity\footnote{CPU activity dataset is available at \url{https://www.openml.org/d/573}.}, and a classification dataset, Satellite image\footnote{Satellite image dataset can be found at \url{https://www.openml.org/d/294}.}. The AE used to find embeddings is the contractive AE. AEs for both datasets have been designed using the same criteria: three hidden layers, the encoding layer having 2 variables and the rest having as much variables as needed so that the compression ratio from the input to the first hidden layer is the same than from the hidden layer to the encoding layer. The resulting architectures are detailed in Fig.~\ref{fig:visualizer}. The AEs have found the projections shown in Fig.~\ref{fig:visualizations}, where the label of each instance is used to color each point. Notice that the AEs have trained without the respective target variables, but there appears to be some degree of separability of classes and different values of the regression variable in each graph.

In order to verify to a certain degree that these embeddings, in addition to producing meaningful visualizations,  contain the necessary information about the data, the mean squared error between each instance and its reconstruction through the AE can be computed. As a reference for comparison purposes, the same reconstruction error can be computed from the two first principal components of the data \revision{and from the encoding found by a basic AE}. Table~\ref{tbl:mse} holds these results, which are very favourable to the contractive AE, since the error is lower in every case. \revision{The difference among both AEs is small, but it serves to deduce that the contractive penalty in the AE does not hinder the reconstruction objective, instead it helps obtain useful low-dimensional embeddings.}

\begin{table}[ht]
    \centering
    \begin{tabular}{lrrrr}
        \toprule
         & \multicolumn{4}{c}{Mean squared error} \\
            \cmidrule(r){2-5}
             & \multicolumn{2}{c}{CPU Activity} & \multicolumn{2}{c}{Satellite} \\
            \cmidrule(r){2-5}
            Method & train & test & train & test \\
        \midrule
            PCA & 0.5577 & 0.5097 & 0.1475 & 0.1483 \\
            Basic AE & \revision{0.5238} & \revision{0.4729} & \revision{0.1136} & \revision{0.1160} \\
            Contractive AE & \revision{\textbf{0.5053}} & \revision{\textbf{0.4546}} & \revision{\textbf{0.1132}} & \revision{\textbf{0.1157}} \\
        \bottomrule               
    \end{tabular}
    \caption{\label{tbl:mse}Mean squared error comparison between the reconstructions of a contractive AE with a 2-variable encoding and the projections to the original feature space from the two principal components of the data. Lower values are better.}
\end{table}

\subsection{Noise reduction}

Similar to searching for interesting representations of data in the encodings of an AE, we can look for a reconstruction that adds value to the input data. One way an AE can help with this is to remove noise from its inputs. This is especially useful when dealing with images~\cite{xie}, sound~\cite{speech} and other kinds of signals~\cite{ecg}, since capture methods usually may introduce some noise and it would be desirable to have a clearer and sharper output.

In general, an AE can be trained to be resilient to input perturbations with a  mere random additive noise at the input. Throughout the optimization stage, the AE only takes as input partially corrupted versions of the training examples and attempts to reconstruct the original ones. Once trained, this AE does not necessarily expect more noisy data, but instead it will have learned to be robust against small changes in its inputs. This type of AE is usually called a denoising AE, and performs well in many scenarios that do not necessarily involve treatment of noisy data~\cite{sdae}.

In this case, nonetheless, the goal is to eliminate potential perturbations in the inputs. Unlike a generic noise reduction filter, which will perform similar operations no matter what data it receives, a denoising AE can be fitted to a specific training set and may thus be more reliable with different kinds of data. More formally, we consider a noise function $\nu$, which generates the corrupted data that the autoencoder trains with to minimize \[\sum_{x\revision{\in\mathcal X}} d(x, g(f(\nu(x))))~.\] The following are some possible noise functions that may be applied:

\begin{itemize}
    \item $\nu(x)=x+\xi$ where $\xi$ is sampled from a Gaussian distribution with small variance
    \item $\nu(x)=x+\xi'$ where $\xi'$ is sampled from a Cauchy distribution with small scale
    \item $\nu(x)=\begin{cases}
        0&\mbox{with low probability}\\x&\mbox{otherwise}
    \end{cases}$
    \item $\nu(x)=\begin{cases}
        0&\mbox{with low probability}\\1&\mbox{with low probability}\\x&\mbox{otherwise}
    \end{cases}$
\end{itemize}

Notice that the Gaussian and Cauchy distributions will usually induce small changes to most inputs, while the zero and zero-one noises will leave most values intact but the change in the corrupted ones will be more drastic. Thus, for a given application, a specific type of corruption function can be selected so that it fits best to the types of noise the samples could have.

\begin{figure}[ht]
    \centering\small
     \resizebox{\linewidth}{!}{
\begin{tikzpicture}[node distance=0.5cm]
\node[layer] (x) {\textsc{input}
\nodepart{second} STL10 + noise
\nodepart{third} $96\times 96\times 3$};
\node[layer, right=of x] (h1) {\textsc{}
\nodepart{second} Conv ($5\times 5$)
\nodepart{third} $96\times 96\times 64$};
\node[layer, right=of h1] (h2) {\textsc{}
\nodepart{second} Conv ($1\times 1$)
\nodepart{third} $96\times 96\times 128$};
\node[layer, right=of h2] (h3) {\textsc{Encoding}
\nodepart{second} Max pooling ($2\times 2$)
\nodepart{third} $48\times 48\times 128$};
\node[layer, below=of h3] (h5) {\textsc{}
\nodepart{second} Deconv ($5\times 5$)
\nodepart{third} $48\times 48\times 64$};
\node[layer, left=of h5] (h6) {\textsc{}
\nodepart{second} Upsampling ($2\times 2$)
\nodepart{third} $96\times 96\times 64$};
\node[layer, left=of h6] (y) {\textsc{output}
\nodepart{second}  Deconv ($3\times 3$)
\nodepart{third} $96\times 96\times 3$};
\draw[-Latex] (x.east) to (h1.west);
\draw[-Latex] (h1.east) to (h2.west);
\draw[-Latex] (h2.east) to (h3.west);
\draw[-Latex] (h3.south) to (h5.north);
\draw[-Latex] (h5.west) to (h6.east);
\draw[-Latex] (h6.west) to (y.east);
\end{tikzpicture}
}
    \caption{Denoising AE architecture for noise reduction}
    \label{fig:graph-denoising}
\end{figure}
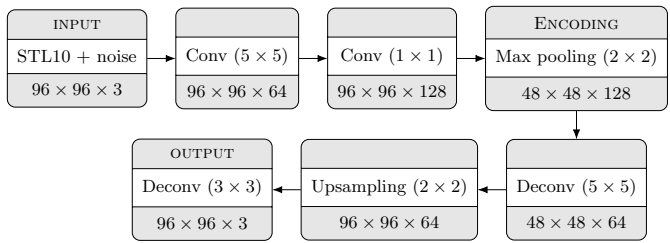

When using denoising AEs, it is also convenient to adapt the type of layers used to the kind of data. For instance, a convolutional AE would be best for noisy images, and an LSTM AE for corrupted signals or sequences. Fig.~\ref{fig:graph-denoising} details a possible encoder-decoder structure for a denoising AE which uses convolutional layers in the encoding phase as well as deconvolution operations during decodification.

When this AE is trained with data from the STL10 dataset~\cite{stl10}, a subset of the ImageNet dataset, the objective function will force it to configure its weights so that input noise is reduced along the network. The noise used in this case has been zeros with a probability of 0.1. The test images measure 96x96 pixels and have also been corrupted with around 10\% of noisy values, which can affect any color channel, so each pixel has a 30\% likelihood of having any of its 3 values altered. The AE was trained during 10 epochs with the training data using optimizer Adam. 

\begin{figure}[ht]
    \centering
    \includegraphics[width=\linewidth,trim={768px 0 1536px 0},clip]{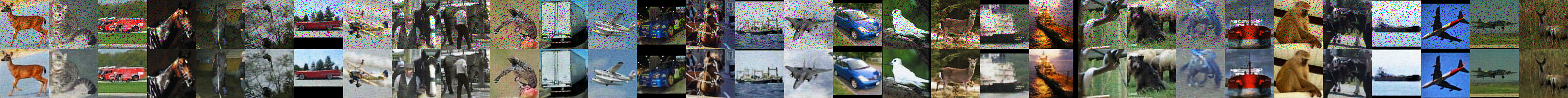}

    \includegraphics[width=\linewidth,trim={1536px 0 768px 0},clip]{denoising-predictions.png}
    \caption{\label{fig:denoised}Random selection of test examples (first and third rows) and their reconstructions (second and fourth rows) via forward passes through the denoising AE.}
\end{figure}

The results can be analyzed in Table~\ref{tbl:noisered}, which shows the designed AE achieves a reduction in the mean squared error of about 89\%. For comparison purposes, a basic AE has also been trained with  Fig.~\ref{fig:denoised} displays some of the test inputs together with their reconstruction by the network. The resulting reconstructions remove most of the noise and appear slightly softer than the originals. 

\begin{table}
    \centering
    \begin{tabular}{lrr}
        \toprule
            Images & Mean squared error & Noise reduction \\
        \midrule
        Reference & 0 & 100\% \\ 
        Noisy & 1656.08  $\pm$ 696.31 & 0\%\\
        \revision{Basic AE}  &  \revision{576.68 $\pm$ 156.53} & \revision{62.14\% $\pm$ 9.54}\\
        Denoising AE &  \revision{\textbf{159.74}} $\pm$ 74.55 & \revision{\textbf{88.94\%}} $\pm$ 6.38\\
        \bottomrule
    \end{tabular}

    \caption{\label{tbl:noisered}Summary of results for noise reduction (average values and standard deviations are provided). Original images without noise are the reference for measuring the mean squared error, and the noise reduction is computed for each image as the percentage decrease in this error. Images are represented by their RGB values from 0 to 255.}
\end{table}

\subsection{Semantic hashing}

Hashing usually refers to the process of summarizing large batches of data in smaller or simpler codes. Hashes are employed in data structures for fast search times, they can be used to find duplicates and to protect data against corruption and manipulation.

This task in particular, semantic hashing~\cite{salakhutdinov}, involves finding binary codes which form buckets of similar data, i.e. when two data points are similar to each other, there is high probability that they will be assigned the same hash. Furthermore, if two similar data points are not hashed identically, their hashes will likely differ in only a few digits. In consequence, a way of finding instances similar to a query instance is to hash it and look for those whose hashes are the same or almost identical. This is the opposite of cryptographic hashing~\cite{katz2014introduction}, where the likelihood of two similar entries obtaining the same hash is almost zero and there is no way of retrieving a document from its hash.

The idea of finding semantic relations between data points is especially useful in document searches: if a query document is provided, then the search method should find those documents in the dataset which match as closely as possible. It is also of application in an image domain, where finding matching binary sequences is much more efficient than comparing two pictures~\cite{carreira2015}.

The approach described in~\cite{salakhutdinov} uses a very simple AE architecture, with an added noise generator after the encoding which forces the encoder to polarize its outputs.

\begin{figure}[ht]
    \centering\small
     \resizebox{\linewidth}{!}{
\begin{tikzpicture}[node distance=0.5cm]
\node[layer] (x) {\textsc{input}
\nodepart{second} Bibtex
\nodepart{third} 1836};
\node[layer, right=of x] (h1) {\textsc{}
\nodepart{second} Dense
\nodepart{third} 512};
\node[layer, right=of h1] (h3) {\textsc{encoding}
\nodepart{second} Dense
\nodepart{third} 7};
\node[layer, right=of h3] (h2) {\textsc{regularizer}
\nodepart{second} GaussianNoise
\nodepart{third} 7};
\node[layer, below=of h1] (h4) {\textsc{}
\nodepart{second} Dense
\nodepart{third} 512};
\node[layer, below=of x] (y) {\textsc{output}
\nodepart{second} Dense
\nodepart{third} 1836};
\draw[-Latex] (x.east) to (h1.west);
\draw[-Latex] (h1.east) to (h3.west);
\draw[-Latex] (h3.east) to (h2.west);
\draw[-Latex] (h2.south) to[out=-90,in=0] (h4.east);
\draw[-Latex] (h4.west) to (y.east);
\end{tikzpicture}
}
    \caption{AE architecture for semantic hashing}
    \label{fig:semantic}
\end{figure}
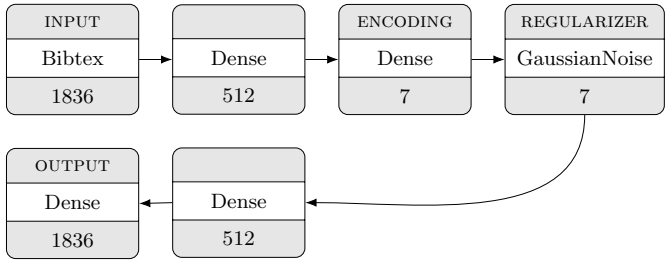

In this case, the Bibtex dataset~\cite{bibtex} was selected to illustrate the application. Fig.~\ref{fig:semantic} shows the AE architecture that was defined for this purpose. The input data provides 1836 binary features which are then projected onto a smaller feature space and lastly onto a 7-dimensional encoding, which is in turn slightly corrupted before decoding. The noise introduced in the encoding during training requires it to take extreme values, for the noise not to affect the reconstruction.

\begin{figure}[ht!]
    \centering
    \includegraphics[width=\linewidth]{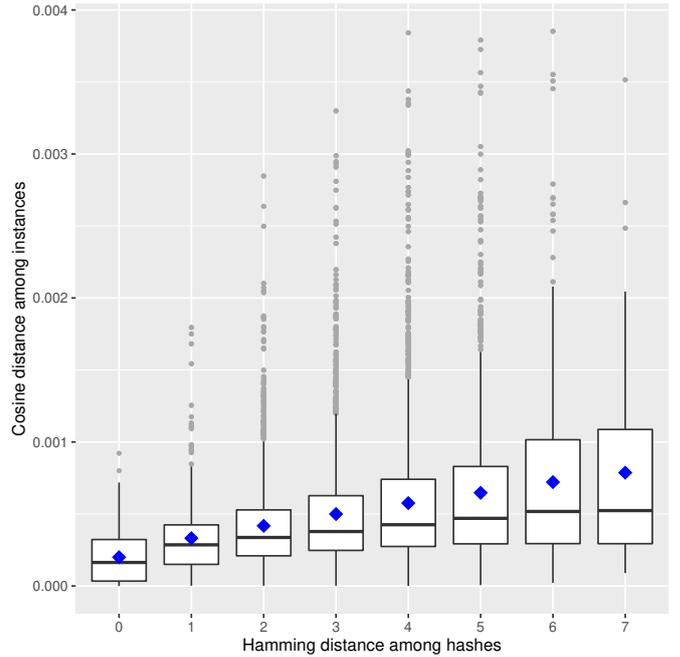}
    \caption{\label{fig:hashingplot}Intercluster cosine distance boxplot according to the hamming distance between hashes. Blue diamonds indicate the mean cosine distance among all pairs of clusters that differ in $k$ digits where $k$ is a Hamming distance. Gray dots indicate outlier cosine distances.}
\end{figure}

In order to assess whether the trained model serves the purpose of semantic hashing, we can group all possible pairs of hashes according to their Hamming distance (e.g. 0001000 and 001001 are 1 digit away from each other, while 1010101 and 0101010 are separated by a Hamming distance of 7). Then, we measure the intercluster distance between those pairs of hashes, computed as the mean cosine distance from each instance in the first cluster to each one in the second. Assuming the clusters group similar instances, the intercluster distance should increase along with the Hamming distance. The distances for this example are illustrated in Fig.~\ref{fig:hashingplot}, which indeed shows simultaneous growth of both.

In addition to quantitatively evaluating the quality of the model, it can be qualitatively analyzed in order to verify whether semantic hashing indeed groups topics in similar hashes. One way of doing this is computing the term frequency-inverse document frequency index (tf-idf)~\cite{tfidf} of the words for each cluster. This way, words that are frequent within a cluster but uncommon along the rest of the test set are considered the most relevant words. Table~\ref{tab:semhashwords} shows a truncated list of hashes used by the AE to cluster documents, along with their most relevant words ranked by tf-idf.

\begin{table}[ht]
    \resizebox{\linewidth}{!}{%
\begin{tabular}{rl}
\toprule
Hash & Relevant words \\
\midrule
0000001  &  \makecell[tl]{thermodynamic, transitions, induced, generalized,\\ completely, interacting} \\
0000011  &  relaxation, barrier, mainly, contribute, surfaces, rights \\
0000010  &  lipoproteins, capacity, oxidation, apo, receptor, recognized \\
0000110  &  identifying, amino, united, capable, matrix, region \\
0000111  &  carbon, storage, enzymes, assessed, notes, roles \\
0000101  &  \makecell[tl]{infrastructure, configuration, challenge, location,\\qualitative, improvement} \\
0000100  &  \makecell[tl]{innovation, construction, ontologies, communities, 1999,\\ located} \\
0001100  &  mining, advances, bioinformatics, er, solved, intelligence \\
0001101  &  reuse, object, perspectives, intelligent, notes, logic \\
0001111  &  trans, reading, behavioral, cultural, 1997, gap \\
0001110  &  ss, siamese, betta, splendens, male, fighting \\
0001010  &  siamese, ss, fighting, male, display, fish \\
0001011  &  treated, barrier, combines, electrostatic, solvent, molecule \\
0001001  &  thermal, boltzmann, origin, bulk, fluctuations, disorder \\
0001000  &  numerically, temperatures, exact, magnetic, glass, zero \\
\bottomrule
\end{tabular}
}
\caption{\label{tab:semhashwords} The first 15 hashes used as semantic codes for clusters found by the AE, ordered in Gray code. The most relevant words are selected according to tf-idf computed for each cluster. They show some common topics between hashes 0001110 and 0001010, and between 0000001, 0001001 and 0001000.}
\end{table}

\subsection{Anomaly detection}\label{sec:anomaly}

Sometimes the objective of a machine learning task is to find unusual behaviors or abnormalities in data, for example, detecting a possible security attack by analyzing server logs, or identifying rare patterns in medical checks. This is known as anomaly detection because the cases of interest are few in contrast to the amount of normal instances, and even in some cases there are no anomalies to train with. In this situation, a traditional classifier cannot solve the problem since it will not be able to assign a class it has not seen before.

An approach to anomaly detection without previously observed anomalous cases is to model those considered typical, and mark as anomalies those instances which do not fit the model. An AE can be used for this purpose, since it can be trained to accurately encode and reconstruct instances following a certain distribution. When the AE is feeded new instances, it is assumed that reconstruction of anomalous data will not be as accurate, since it should follow a different distribution~\cite{petsche, sakurada, park}. More formally, the hypothesis of this methodology is that, when trained with normal data, $d(x, g(f(x)))$ will be very small when $x$ is normal and very high when $x$ is anomalous.

An useful application of anomaly detection where real world data will generally lack anomalies is network intrusion \cite{mirsky5kitsune,dlnid}, that is, the detection of potential security attacks and malicious accesses to a server. The straightforward approach is to continuously log server accesses, and extract data from a period of time where usage has been normal. By means of these data, an AE can be trained to recognize typical usage parameters. Then, new log accesses are constantly feeded to the AE in order to predict their reconstruction error. In the case that several successive errors are much higher than the mean, an attack may be underway.

The AE used for this purpose will work as follows: the encoding layer will perform a drastic dimensionality reduction in order for it to model the most essential information from the training data, which does not include any anomaly. This should help have low error rates on normal data, similar to training instances, but very high ones on anomalous data. In general, this may not work well for uncommon, isolated anomalies, but it is useful when anomalies are several in sequence, so this strategy is especially designed for time series data.

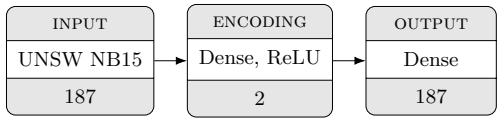
\begin{figure}[ht]
    \centering\small
     \resizebox{0.75\linewidth}{!}{
\begin{tikzpicture}[node distance=0.5cm]
\node[layer] (x) {\textsc{input}
\nodepart{second} UNSW NB15
\nodepart{third} 187};
\node[layer, right=of x] (h3) {\textsc{encoding}
\nodepart{second} Dense, ReLU
\nodepart{third} 2};
\node[layer, right=of h3] (y) {\textsc{output}
\nodepart{second} Dense
\nodepart{third} 187};
\draw[-Latex] (x.east) to (h3.west);
\draw[-Latex] (h3.east) to (y.west);
\end{tikzpicture}
}
    \caption{Denoising AE architecture for anomaly detection}
    \label{fig:anomaly}
\end{figure}

The dataset treated in this example is UNSW-NB15~\cite{unsw}, which has 3 nominal variables and 42 numerical descriptors. Since AEs cannot work directly with nominal variables, these have been converted into dummy binary variables. In addition, any anomalous data from the training subset has been removed. In total, 37000 instances with 187 features  are being introduced as the training input of the AE, whose architecture is shown in \ref{fig:anomaly}. The extraction of two features is sufficient to model an approximation of most of the normal data, but cannot preserve enough information for the reconstruction of most anomalies. 

\begin{figure}[ht]
    \centering
    \includegraphics[width=.8\linewidth]{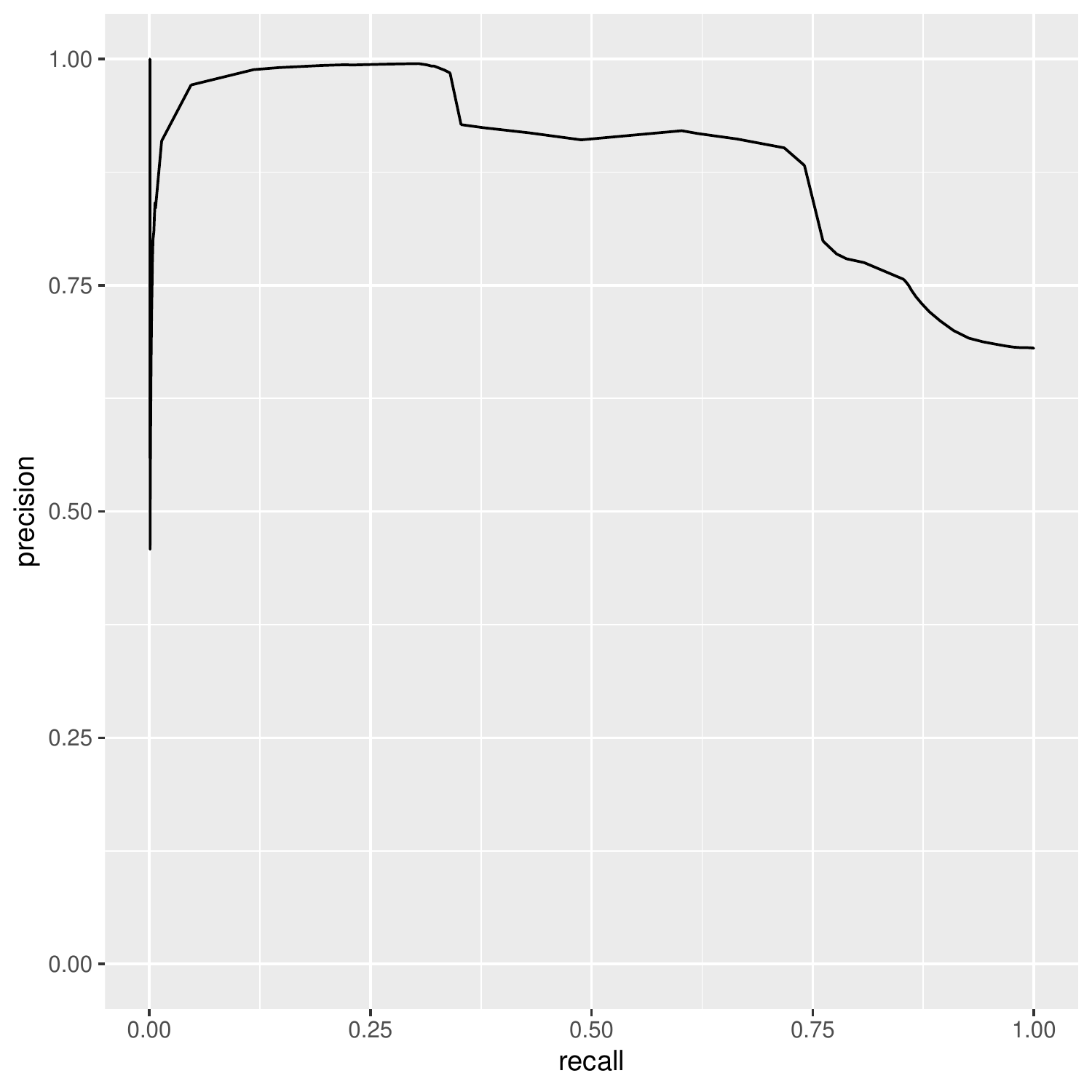}
    \caption{Precision-recall curve for the detection of individual anomalies in the UNSW dataset.}
    \label{fig:anomaly-pr}
\end{figure}

\begin{figure*}[ht!]
    \centering
    \includegraphics[width=.5\linewidth]{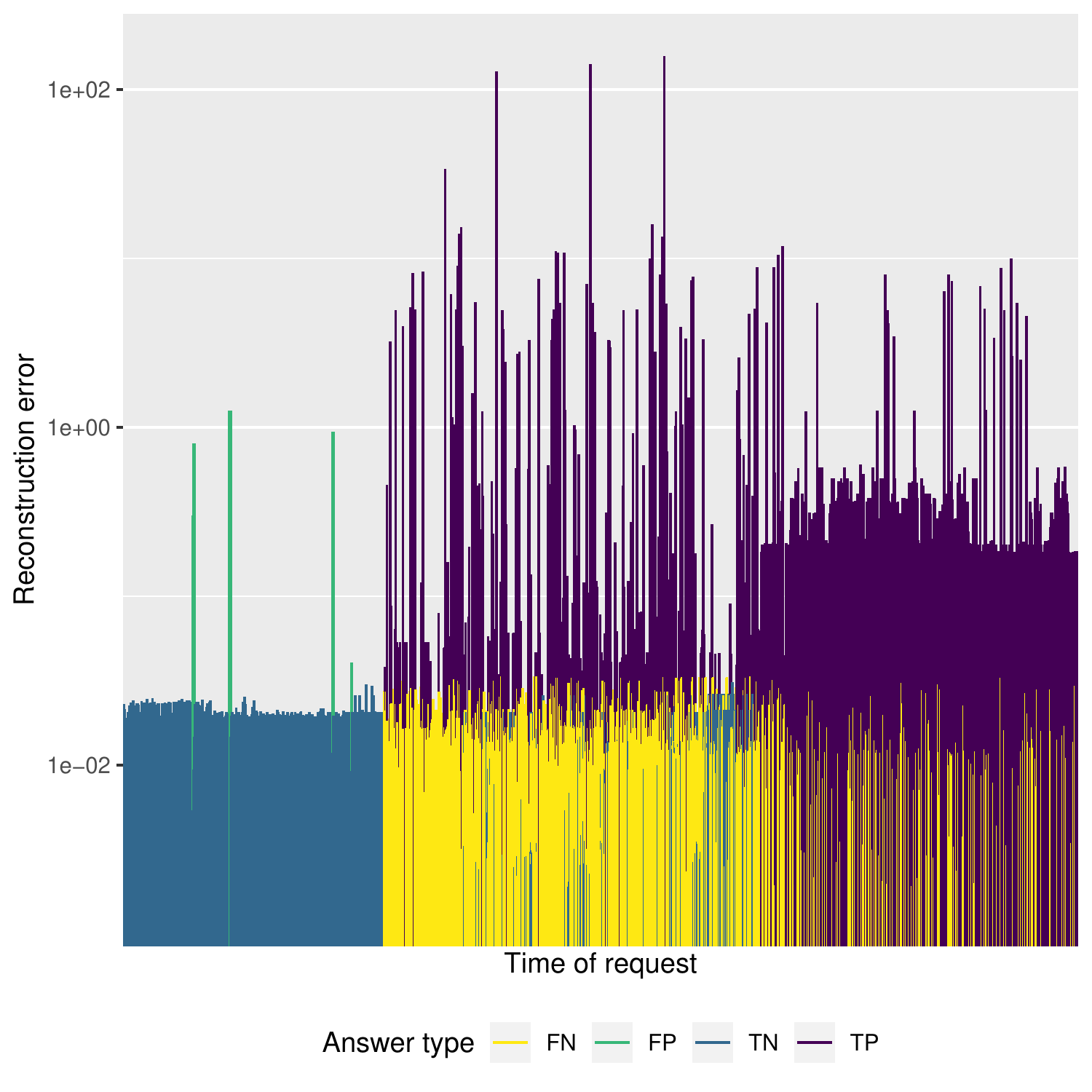}%
    \includegraphics[width=.5\linewidth]{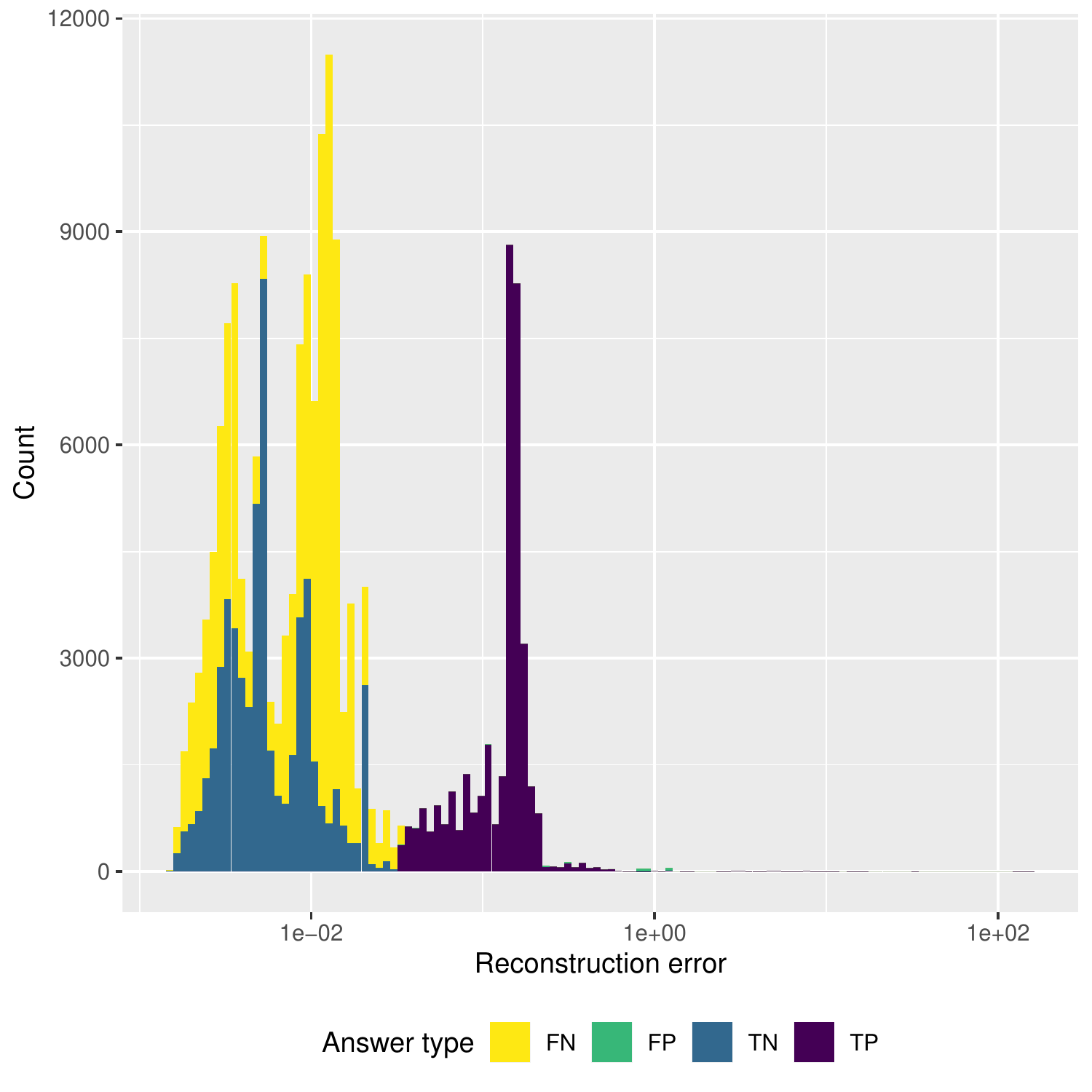}
    
    \caption{\label{fig.anomaly-hist}Reconstruction error of the AE during test. The graph on the left shows the reconstruction error of each request in sequence, where the detection threshold is set to the mean training error plus 6 times its standard deviation. The histogram on the right shows the amount of hits and misses according to the reconstruction error.}
\end{figure*}

The results of training this model are summarized in Fig.~\ref{fig:anomaly-pr} and Fig. \ref{fig.anomaly-hist}. The first is a precision-recall curve which gives details about the fraction of detections which are actually anomalies and the ratio of detected anomalies among all of them. We find that it is possible to detect more than half the anomalies without obtaining too many false alarms. Since the test dataset contains many more anomalies than normal instances and the objective is to detect abnormal sections more than to find every individual anomaly, a recall of around 50\% could be enough as long as the precision is high so that few false alarms are raised.

Indeed, Fig.~\ref{fig.anomaly-hist} graphs the reconstruction error for each test instance and shows that when an adequate threshold is chosen, anomalous sections can be easily detected with very few isolated false alarms that can be discarded. In this case, the chosen threshold is the mean reconstruction error plus 6 times its standard deviation, but it could be tuned high or low in order to adjust the sensitivity of the detection.

\subsection{Instance generation}

The representation learned by an AE may be useful to encode or reconstruct individual instances from a training set, but in certain cases it will be very convenient to ensure that this representation is actually attempting to perform some kind of manifold learning, mapping the feature space onto a smaller space in a way that makes sense to work with the whole encoding space. This encoding space would allow to predict a reconstruction for encodings that do not come from an instance in the original feature space, and still produce a coherent result. For instance, an useful application would be to generate new images of faces similar to those in a training set but not identical to any of them. This is usually harder to achieve with simple operations such as interpolation, because they would compute many images that do not represent faces.

There are several variants of AEs that can fulfill this purpose, namely variational~\cite{variational}, adversarial and contractive AEs. Variational as well as adversarial AEs force a prior distribution in the encodings in different ways, which allows to sample new instances by taking points from this space and projecting them onto the original feature space via reconstruction ($g$). The contractive AE, on the contrary, only imposes a regularization which promotes instances to be mapped to encodings near their neighbors. This helps the autoencoder perform transformations that find manifolds in the data, since local structure is preserved. The manifold can then be traversed in order for the decoder to generate new instances.

Variational AEs are stochastic in the sense that they do not map each instance to a single point in the embedding space, but a distribution instead. This is usually a normal distribution, defined by its mean and standard deviation. Then, a reconstruction is produced by sampling that distribution and propagating the results through the decoder network. The objective function in this AE combines the clustering behavior of the reconstruction loss function with a regularization loss which forces the distribution to be as similar as possible to, generally, a multivariate unit Gaussian. This helps the AE extract a very compact representation which only preserves the necessary information to provide a reconstruction of the input.

\begin{figure}[ht]
    \centering\small
    \resizebox{\linewidth}{!}{%
\begin{tikzpicture}[node distance=0.5cm]
\node[layer] (x) {\textsc{input}
\nodepart{second} AT\&T faces
\nodepart{third} $ 64 \times 64\times 1 $};
\node[layer, right=of x] (h1) {\textsc{}
\nodepart{second} Conv ($3 \times 3$)
\nodepart{third} $ 32 \times 32 \times 8 $};
\node[layer, right=of h1] (h2) {\textsc{}
\nodepart{second} Conv ($3 \times 3$)
\nodepart{third} $ 16 \times 16 \times 16 $};
\node[layer, right=of h2] (h3) {\textsc{}
\nodepart{second} Conv ($3 \times 3$)
\nodepart{third} $ 8 \times 8 \times 32 $};
\node[layer, below=of h3] (h4) {\textsc{mean \& var}
\nodepart{second} Dense
\nodepart{third} $ 32 + 32 $};
\node[layer, left=of h4] (h5) {\textsc{sampling}
\nodepart{second} Custom
\nodepart{third} $ 32 $};
\node[layer, left=of h5] (h6) {\textsc{}
\nodepart{second} Dense
\nodepart{third} $ 8 \times 8 \times 8 $};
\node[layer, left=of h6] (h7) {\textsc{}
\nodepart{second} Deconv ($3 \times 3$)
\nodepart{third} $ 16 \times 16 \times 32 $};
\node[layer, below=of h7] (h8) {\textsc{}
\nodepart{second} Deconv ($3 \times 3$)
\nodepart{third} $ 32 \times 32 \times 16 $};
\node[layer, right=of h8] (h9) {\textsc{}
\nodepart{second} Deconv ($3 \times 3$)
\nodepart{third} $ 64 \times 64 \times 8 $};
\node[layer, right=of h9] (y) {\textsc{output}
\nodepart{second} Deconv ($3 \times 3$)
\nodepart{third} $ 64 \times 64 \times 1 $};
\draw[-Latex] (x.east) to (h1.west);
\draw[-Latex] (h1.east) to (h2.west);
\draw[-Latex] (h2.east) to (h3.west);
\draw[-Latex] (h3.south) to (h4.north);
\draw[-Latex] (h4.west) to (h5.east);
\draw[-Latex] (h5.west) to (h6.east);
\draw[-Latex] (h6.west) to (h7.east);
\draw[-Latex] (h7.south) to (h8.north);
\draw[-Latex] (h8.east) to (h9.west);
\draw[-Latex] (h9.east) to (y.west);
\end{tikzpicture}
}
    \caption{Variational AE architecture for instance generation. The sampling layer draws a sample from the vector of normal distributions with means and variances given by the previous layer.}
    \label{fig:variational}
\end{figure}
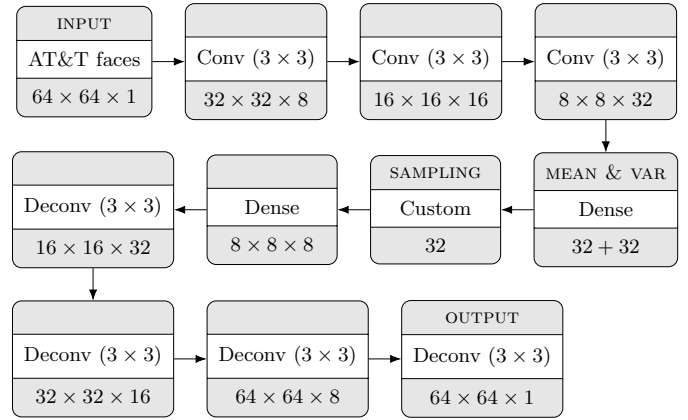

\begin{figure}[ht!]
    \centering
    \includegraphics[width=\linewidth]{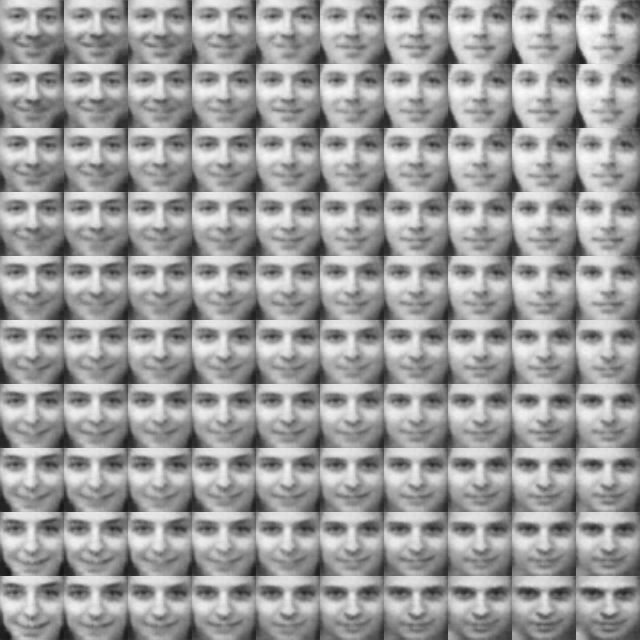}
    \caption{\label{fig:sampledfaces}Faces sampled from the encoding space of a variational AE, using interpolations between the projections of images in the original dataset}
\end{figure}

In this example, a variational AE following the structure in Fig.~\ref{fig:variational} is trained to generate human faces that do not belong to any person, since they will not be present in the training dataset. The input data used during training belong to the AT\&T faces dataset\footnote{AT\&T faces dataset is available at \url{https://www.openml.org/d/41083}.}, also known as Olivetti faces dataset. The resulting model can be sampled by feeding arbitrary values to the generator component, which then outputs previously unseen images. Fig.~\ref{fig:sampledfaces} shows some representative examples of the generated faces using this AE.

\subsection{\revision{Other applications}}\label{sec:oth}

Apart from the previous selection of applications approached with representation learning techniques based on AEs, there are many other situations where AEs can be applied to extract features from data. The following are learning applications present in the literature that fell out of the scope of this article.

\subsubsection{Image superresolution} This problem consists in building a high resolution image from a low resolution sample, such as a thumbnail. By using an AE trained with low resolution images and another with the high resolution ones, a map can be trained from the first encoding to the second~\cite{superresolution}. This way, the encoder from the first AE can be connected to the decoder from the second AE and the resulting network can be fine-tuned. During prediction it suffices with feeding a low resolution image through the new network, which will encode it and decode it through the high resolution decoder, producing a higher quality image.

\subsubsection{Image compression} Images are usually compressed with algorithms designed for this specific purpose, e.g. the JPEG standard~\cite{jpeg}. Since a compression mechanism must include a component which compresses the image and another which performs decompression, AEs can be trained in different ways to treat this problem as well~\cite{balle2016end,compression,cheng2018deep}, even surpassing the capacity of JPEG2000 especially at low bit rates.

\subsubsection{Transfer learning} In a transfer learning task, the learner must make use of the knowledge extracted from data in a given domain to apply it to a different domain. This may consist in using pre-trained networks with a large dataset to use them with a small dataset by a fine-tuning process. However, when labels for the large dataset are not available, the first stage will necessarily be unsupervised~\cite{bengio2012deep}, in which case an AE can be trained and its extracted features can initialize a network for a supervised problem with a dataset from other domain.

\subsubsection{Human pose \revision{and facial features}} Human pose recovery is an application specific to image and video data where people appear and the aim is to recognize the pose of each person from the visual information, i.e., to generate a skeleton describing the position and orientation of the legs, arms and the rest of the body. One of the challenges is to model this skeleton as a 3D object while images are only 2D. AEs have been used as the core of a human pose recovery model~\cite{hong2015multimodal} for extracting an inner pose representation which then maps onto a representation of the 3D pose and is decoded as a 3D pose. This process is, in fact, achieved with two AEs, one for each inner representation required, which are then connected through the representation mapping. \revision{In a similar way, facial expression recognition aims to identify the human emotional state from facial images. An approach based on deep sparse autoencoders \cite{facialexpr}, which are used to extract robust and discriminative features, has been developed to tackle this task.}

\subsubsection{3D shape learning} Extracting features from three-dimen\-sional shapes usually has a high computational cost but it is fundamental for tasks such as 3D object retrieval and matching. There are several AE-based models for automatic feature extraction that can help model this type of data~\cite{3d1,3d2,3d3}. These range from simple stacked AEs to combinations of convolutional AEs and extreme learning machines. In general, retrieving similar objects to a given input consists in encoding the input and comparing the result to the codes of known objects in order to find the nearest or most similar ones.

\subsubsection{Recommender systems and tagging systems} Recommender systems are filters that seek to predict user preferences for products, taking into account previous choices or ratings. Collaborative filters for recommendation combine the information of different users to build predictions. In~\cite{li2017collaborative}, a collaborative variational AE for recommendation is developed. It models the implicit relationships among items and users by making use of a shared latent representation and the variational regularization. A task similar to recommendation is tagging, since tags can be ranked for an item according to its similarity to other items. AEs have been also used as the core of tagging systems~\cite{wang2015relational} using denoising AEs and relational denoising AEs.

{\revisioncolor
\section{Challenges for autoencoder progress and prospects on explainability}\label{sec:ch}

Along this section, several difficulties and consequences of using AEs in machine learning are explored. Some brief comments are provided beforehand on the current state of explainability and transparency in artificial intelligence (AI), in order to understand how they could affect the way AEs are designed and used.

First, we introduce the most popular approaches to finding transparent and explainable machine learning models. Later, we develop on the ways AEs can help build interpretable solutions to different problems, by learning disentangled and fair features.

\subsection{State and prospects on explainability}

Explainable AI \cite{arrieta2020explainable} encompasses many concepts around the idea that people should be able to understand how trained machine learning models work and why they make their decisions.

The recent surge in interest in explainable models derives from the bias found in existing models as well as the search for AI safety \cite{aisafety}. The first issue involves models that make decisions potentially affecting human beings and those decisions can discriminate against certain population groups, e.g. people of color or women. For instance, a prediction model for criminal recidivism was found to be heavily biased against African-American people \cite{angwin2016machine}. The second concept relates to the presence of relatively autonomous agents, such as robots, which execute the actions computed by a machine learning model. These models sometimes find unexpected ways to optimize their reward function (reward hacking) \cite{bird2002evolved}, even without completing the objective or having other potential consequences (side effects).

\subsubsection{Model transparency}

The issues above reflect the fact that we should not completely trust trained models unless we can comprehend the ways they are making decisions and predictions. This has attracted the interest of researchers, domain experts and users to more explainable models and strategies to explain \textit{black-box} models, a category which includes most deep learning techniques.

The variety of algorithms to fit machine learning models to data presents a tradeoff between performance and explainability: usually, a simpler, more explainable model is less performant than an opaque model. As a consequence most simple models, such as decision trees, rule-based learners and k-nearest neighbors, are considered transparent, since they provide an interpretable behavior out of the box.

When a model is not transparent enough, there are two main ways to approach explainability: one can use different post-hoc explainability approaches, or modify the model to facilitate our understanding of its decision process. Some new models derived from deep feature learners are designed to improve transparency and be more self-explanatory. 

One way deep learning models can increase transparency is by highlighting which input features are causing their predictions. For example, attention-based models \cite{cao2015look} have an embedded scoring technique which highlights the zones in the input that are being taken into account to make predictions. This works for image classification and object detection \cite{wang2017residual} as well as for document processing~\cite{yang2016hierarchical}. 

A different proposal for transparent image classification is a convolutional neural network-based classifier which identifies prototypes in similar images \cite{chen2019looks}, that is, it provides examples on images of the same class that justify the prediction.

\subsubsection{Explainability techniques}

When an opaque model is used, there are still ways to improve our understanding of its inner workings or its predictions. In many cases, a post-hoc explainability technique may be applied. 
The different methods that can render a model more interpretable are usually categorized into two groups. They can be either model-agnostic, if they work independently of the model used, or model-specific, otherwise. 

Some examples of model-agnostic approaches and tools are the following:

\begin{itemize}
    \item Local approximations. LIME \cite{ribeiro2016should} this is a method which linearly performs a local approximation of a classifier or regressor, in a way which is interpretable. An AE-based variant of LIME has been developed to improve its stability \cite{shankaranarayana2019alime}.
    \item FairML \cite{adebayo2016fairml}. The FairML toolbox can find strong dependencies between model outputs and the input features. 
    \item Sensitivity analysis \cite{baehrens2010explain}. It is a computation based on the derivative of the conditional probability of not predicting a class given the input features. This defines a vector field where each vector indicates the direction an instance needs to be moved to, so as to be classified differently.
    \item Auditing. 
    Trained models can be repeatedly tested against different inputs in order to analyze how the outputs are affected. These inputs, however, need to be provided according to some criteria. As a way to compute direct and indirect influence of each feature in the output of a model, there is a procedure which obscures the effect of a variable in the data \cite{adler2018auditing}. It works without retraining the model, and can assess the degree in which a feature is relevant to a classifier. There are several other approaches to analyzing direct influence of a feature in the output of a model~\cite{henelius2014peek,datta2016algorithmic}.
    \item Counterfactuals \cite{molnar2019,wachter2017counterfactual,2020arXiv200311323B}. This is an approach with a similar objective to auditing but from a different perspective. Finding a counterfactual consists in detecting the smallest possible change in feature values that causes an alteration to the prediction of the model. These serve as an explanation for the ``closest possible world'' where the prediction would have been different, without providing further insight into the decision process.

\end{itemize}

There are several specific techniques for explaining the outputs of deep learning models:
\begin{itemize}
\item Layer-wise relevance propagation \cite{bach2015pixel}. This is a methodology for visualization of pixel-wise contributions to predictions, where classifiers are decomposed into several layers of computation, so the relevance of each pixel is found by propagating relevance backwards through the network.
\item Saliency map generation \cite{simonyan2013deep}. Saliency maps are heatmaps where the most relevant features from the input are highlighted. These are usually applied to convolutional neural networks in order to obtain the image regions that cause the output for each instance.
\item DeepLIFT \cite{shrikumar2017learning}. This is a technique for computing relevance for each input feature to a neural network, by assigning contibution scores to each neuron according to its activation given a specific input.
\item SHAP \cite{NIPS2017_7062}. This tool provides several model-specific techniques which find local explanations for different models based on Shapley values from game theory. In particular, it includes DeepExplainer and GradientExplainer, which apply to deep learning models.
\item Traceability \cite{aravantinos2018traceability}. This is a more theoretical concept from the field of software development that could be applied to deep neural models. It seeks to describe how each component of a final inference model is related back to its training model, the dataset, hyperparameters and all the way up to some high level requirements on what task the model should carry out. Being able to trace every item in the development of a deep neural networks to a higher level cause could serve to ensure that all choices such as hyperparameters and architecture are well justified.
\end{itemize}

\subsection{Current challenges and influence in future work}

As discussed in the previous section, most of the well-known explainability techniques involve analysis of features in one way or another. The contribution that AEs can provide in this field is, therefore, substantial. This is due to the fact that AEs can transform a set of highly dependent, correlated features in a different set of independent, interpretable ones, by using adequate regularizations. In this section, we comment on different ways to learn features that are \textit{meaningful} and \textit{fair}, and on recent developments for also improving the explainability of the feature extraction process itself.

\subsubsection{Improving features: disentanglement and fairness}

One way extracted features can improve their quality is by holding an understandable meaning by themselves, e.g. a model could train with face pictures and extract a feature for hair color, another one for nose size, etc. These new features would be much more useful than the original ones which represent individual pixels. This task is usually known as \textit{feature disentanglement}.

Some recent AE models whose objective is to disentangle features are Total Correlation VAE \cite{chen2018isolating}, Wasserstein AE \cite{rubenstein2018learning} and InfoGAN \cite{chen2016infogan}. All of these are generative models, so, as a result, extracted features not only provide interpretable meaning to instances, but can also be sampled in order to generate unseen examples in a way that resembles the manipulation of existing instances: for example, a model could generate a realistic face similar to an existing image but changing blonde hair to black.

Another step forward in improving learned representations is forcing these to become fair \cite{zemel2013learning}, which means that the extracted features obfuscate information about membership to potentially discriminated groups, e.g. gender or ethnicity. Fairness usually applies only in contexts where model predictions affect human lives, e.g. job applications, legal proceedings, etc. A statistic can be defined to measure the discrimination of a classifier with respect to a binary variable. The objective is then to optimize a tradeoff between classification accuracy and discrimination~\cite{edwards2015censoring}.

There already exist AE-based models for learning fair representations. In \cite{madras2018learning}, an adversarial AE-based classifier is proposed where the adversary attempts to predict the sensitive (potentially discriminatory) attributes from the encoding, but its prediction ability is minimized by the AE and classifier. The objective function can be adjusted according to the desired type of fairness. Another model in~\cite{creager2019flexibly} consists in a variational AE which disentangles sensitive information from the non-sensitive latent features and is flexible in the sense that potentially sensitive information can be retained or removed from the encoding during inference.

\subsubsection{Explainable feature learning}

The described approaches provide the possibility of explaining the end predictions of other models, as well as rendering them fairer. However, as has been extensely discussed in this work, the extracted features can be the actual core of a solution to many problems. As a consequence, it would be necessary as well to develop strategies which facilitate the explainability of the transformations an AE can perform in order to learn features. This is an area only explored very recently, but there are already some developments.

Variational AEs can be used to detect anomalies, similarly to the denoising AE explained in Section~\ref{sec:anomaly}. In addition, they enable another, more explainable way of detecting anomalies: computing the gradients of the reconstruction error with respect to the inputs \cite{nguyen2019gee}. This allows to notice which input features are contributing to the error, and to cluster anomalies according to this same criterion.

A different approach to improving the explainability of the embedding consists in restricting the operations each neuron performs to just logical AND/OR operators \cite{ALHMOUZ2019104874}, which limits the origin of each extracted feature to a relatively simple logical combination of the input features, thus facilitating its interpretability.

\subsubsection{Influence in future works}

There is currently much to be researched in the area of explainable AEs as well as AEs which help explain other models by extracting better features. The current trends focus especially on generative models such as variational AEs for these purposes, and will probably continue to do so, even if some diversification is achieved as new works appear.

The adaptability of AEs to many different problems, illustrated in previous sections, together with the possibility of producing interpretable and fair features, may lead to an increase in usage of these models throughout all kinds of machine learning applications. 

In our future work, we intend to approach explainable feature learning in the context of AEs, that is, find AE-based models that extract features and at the same time provide an understandable meaning to the mapping from the original features to the encoded ones. Ideally, an explainable feature learner should not be restricted to one end application, but could be used for many purposes, as common AEs already can.
}

\section{Conclusions}\label{sec:co}

Throughout this text, we have summarized the traditional alternatives for learning representations, the origins and essential characteristics of AEs, including how to introduce certain behaviors into the coding layer. 

Later, we have thoroughly examined several case studies of AE applications in unstructured data as well as images and sequences: data visualization, image denoising, semantic hashing, anomaly detection and instance generation. Other applications have also been briefly discussed: image superresolution, image compression, transfer learning, human pose recovery and recommender systems. 

\revision{An introduction to the state of the art in explainable AI and its application to the field of AEs has been provided as well. AEs have notoriously contributed to the areas of feature disentanglement and fair representations, and there have been some recent developments on explainable feature learning as well.}

We can conclude that AEs are a versatile framework for solving a wide variety of problems where a central task is to learn representations of the data. They can adapt to a given problem in structure as well as in the objective they optimize. This way, if the solution to a problem can be modeled with a transformation of the feature space onto another space, there will be many instances where the parameters of the transformation can be learned by an AE.

\textbf{Acknowledgments}: D. Charte is supported by the Spanish Ministry of Science, Innovation and Universities under the FPU National Program (Ref. FPU17/04069). This paper is partially supported by the Spanish National Research Projects TIN2015-68854-R and TIN2017-89517-P and the project DeepSCOP Ayudas
Fundaci{\'o}n BBVA a Equipos de Investigaci{\'o}n Cient{\'i}fica en Big Data
2018. 


\end{document}